\documentclass{article}
\usepackage{geometry}                
\usepackage{graphicx}
\usepackage{amssymb}
\usepackage{epstopdf}
\usepackage{textcomp}
\usepackage{color}
\usepackage{breakurl}
\usepackage[breaklinks]{hyperref}

\usepackage{listings}
\usepackage[square,comma,sort&compress]{natbib}%

\usepackage[toc,section=section]{glossaries}%
\newacronym{aaai}{AAAI}{Association for the Advancement of Artificial Intelligence}
\newacronym{acm}{ACM}{Association for Computing Machinery}
\newacronym{aco}{ACO}{Ant Colony Optimization}
\newacronym{ANOVA}{ANOVA}{Analysis of Variance}
\newacronym{aslib}{ASlib}{Algorithm Selection Library}
\newacronym{bfgs}{BFGS}{Broyden-Fletcher-Goldfarb-Shanno}
\newacronym{bbob}{BBOB}{Black-Box-Optimization-Benchmarking}
\newacronym{CNN}{CNN}{Convolutional Neural Network}
\newacronym{cmaes}{CMA-ES}{Covariance Matrix Adaptation Evolution Strategy}
\newacronym{COCO}{COCO}{Comparing Continuous Optimizers} 
\newacronym{crn}{CRN}{Common Random Numbers}
\newacronym{crs4ea}{CRS4EA}{Chess Rating Systems for Evolutionary Algorithms}
\newacronym{dace}{DACE}{Design and Analysis of Computer Experiments}
\newacronym{DE}{DE}{Differential Evolution} 
\newacronym{doe}{DOE}{Design of Experiments}
\newacronym{ea}{EA}{Evolutionary Algorithm}
\newacronym{ec}{EC}{Evolutionary Computation}
\newacronym{ecdf}{ECDF}{Empirical Cumulative Distribution Function}
\newacronym{eda}{EDA}{Exploratory Data Analysis}
\newacronym{edalgo}{EDAlgo}{Estimation of Distribution Algorithm}
\newacronym{ego}{EGO}{Efficient Global Optimization}
\newacronym{ela}{ELA}{Exploratory Landscape Analysis}
\newacronym{ert}{ERT}{Expected Running Time}
\newacronym{es}{ES}{Evolution Strategy}
\newacronym{fe}{FE}{Function Evaluation}
\newacronym{ga}{GA}{Genetic Algorithm}
\newacronym{gecco}{GECCO}{Genetic and Evolutionary Computation Conference}
\newacronym{GAN}{GAN}{Generative Adversarial Network} 
\newacronym{HAN}{HAN}{Hierarchical Attention Network}
\newacronym{gpd}{GPD}{Generalized Position-Distance}  
\newacronym{ieee}{IEEE}{Institute of Electrical and Electronics Engineers}
\newacronym{irace}{irace}{iterated racing}
\newacronym{LSTM}{LSTM}{Long Short Term Memory}
\newacronym{lhd}{LHD}{Latin Hypercube Design}
\newacronym{macoda}{MACODA}{Many Criteria Optimization and Decision Analysis}
\newacronym{mcesim}{MceSim}{Multi-Car Elevator Simulator}
\newacronym{nflt}{NFLT}{no free lunch theorem}
\newacronym{MBO}{MBO}{Model-Based Optimization} 
\newacronym{MLE}{MLE}{Maximum Likelihood Estimation} 
\newacronym{moo}{MOO}{Multi-Objective Optimization}
\newacronym{nm}{NM}{Nelder Mead}
\newacronym{ofat}{OFAT}{One-factor-at-a-time}
\newacronym{paramils}{ParamILS}{Iterated Local Search in Parameter Configuration Space}
\newacronym{pbo}{PBO}{Pseudo-Boolean Optimization}
\newacronym{pdsc}{pDSC}{practical Deep Statistical Comparison}
\newacronym{PSO}{PSO}{Particle Swarm Optimization} 
\newacronym{RNN}{RNN}{Recurrent Neural Network}
\newacronym{sGAN}{seqGAN}{Sequential GAN}
\newacronym{sat}{SAT}{Boolean Satisfiability}
\newacronym{smac}{SMAC}{Sequential Model-based Algorithm Configuration}
\newacronym{SCDT}{SCDT}{Structured Categorical Decision Trees} 
\newacronym{SPO}{SPO}{Sequential Parameter Optimization}
\newacronym{spot}{SPOT}{Sequential Parameter Optimization Toolbox}
\newacronym{tsp}{TSP}{Traveling Salesperson Problem}

\renewcommand{\cite}{\citep}

\definecolor{mygreen}{rgb}{0,0.6,0}
\definecolor{mygray}{rgb}{0.5,0.5,0.5}
\definecolor{mymauve}{rgb}{0.58,0,0.82}

\usepackage{csquotes}
\usepackage[font=small,labelfont=bf]{caption}

\graphicspath{{figures/}{pictures/}{images/}{./}} 

\title{Simulation of an Elevator Group Control Using Generative Adversarial Networks and Related AI Tools}
\author{Tom Peetz\\%
\and Sebastian Vogt\\ %
\and Martin Zaefferer\\ %
\and Thomas Bartz-Beielstein\\ %
\and
Institute for Data Science, Engineering, and Analytics \\
Technische Hochschule K{\"o}ln\\
Steinm{\"u}llerallee 1,
51643 Gummersbach
}

\begin{document}
\maketitle

\abstract{
Testing new, innovative technologies is a crucial task for safety and acceptance.
But how can new systems be tested if no historical real-world data exist? 
Simulation provides an answer to this important question.
Classical simulation tools such as event-based simulation are well accepted.
But most of these established simulation models require the specification of many parameters. Furthermore, simulation runs, e.g., CFD simulations, are very time consuming.
Generative Adversarial Networks (GANs) are powerful tools for generating new data for a variety of tasks. Currently, their most frequent application domain is image generation.
This article investigates the applicability of GANs for imitating simulations. 
We are comparing the simulation output of a technical system with the output of a GAN. 
To exemplify this approach, a well-known multi-car elevator system simulator was chosen. Our study demonstrates the feasibility of this approach. It also discusses pitfalls and technical problems that occurred during the implementation. Although we were able to show that in principle, GANs can be used as substitutes for expensive simulation runs, we also show that they cannot be used ``out of the box''. Fine tuning is needed. We present a proof-of-concept, which can serve as a starting point for further research. 
} 

\section{Introduction}

Modern high-rise buildings require a lot of space for elevators. 
Classic elevator systems generally use central machine rooms and steel ropes for moving the elevator cars.
Therefore, a shaft can only hold one car or a fixed stack of cars. 
Steel cables also limit the amount of reachable levels for the elevator, because the cables are too heavy~\citep{miyamoto_mcesim_2008}.
Recent approaches suggest elevator cars with independent linear motors, because they do not require any cables.
Multiple cars can be operated independently in every shaft.
These cars need to be controlled more carefully due to possible collisions, which cannot happen in single car shafts.
They have to be controlled efficiently to maximize utility while reducing energy consumption, maintenance costs, and waiting times for passengers.
Modeling such complex systems accurately is a demanding task. 
Additional elevator behavior data could improve the required algorithms for optimization and control.
Simulation software is available, but is limited in features and general simulation accuracy.
Generally, existing simulators focus on specific building and elevator types.
We propose to generate additional data for elevator systems with the application of \glspl{GAN}.
They are promising tools mainly used for generating images. 
Less frequent applications include text and language processing.

The goal of this paper is to explore the possible applications of \glspl{GAN} to create machine readable logs for 
the described elevator systems based on log data from an existing simulator.
As previous work has shown, generative networks such as standalone \glspl{RNN} suffer from issues like exposure bias resulting in the generation of data which are never observed in training data~\cite{bengio_scheduled_nodate}.
\glspl{GAN} offer a solution to this by adding a neural network, the discriminator,  which tries to classify the generated data as ``real''  or ``fake''.

The remainder of this report is structured as follows:
section~\ref{sec:related} presents related approaches. Section~\ref{sec:methods} describes materials and methods. Results from our first experiments are shown in Section~\ref{sec:result} and discussed in Section~\ref{sec:discussion}. 
Further steps are presented in Section~\ref{sec:outlook}. A conclusion is given in Section~\ref{sec:conclusion}.

\section{Related Work}\label{sec:related}

\begin{figure}
    \centering
    \includegraphics[width=0.5\textwidth]{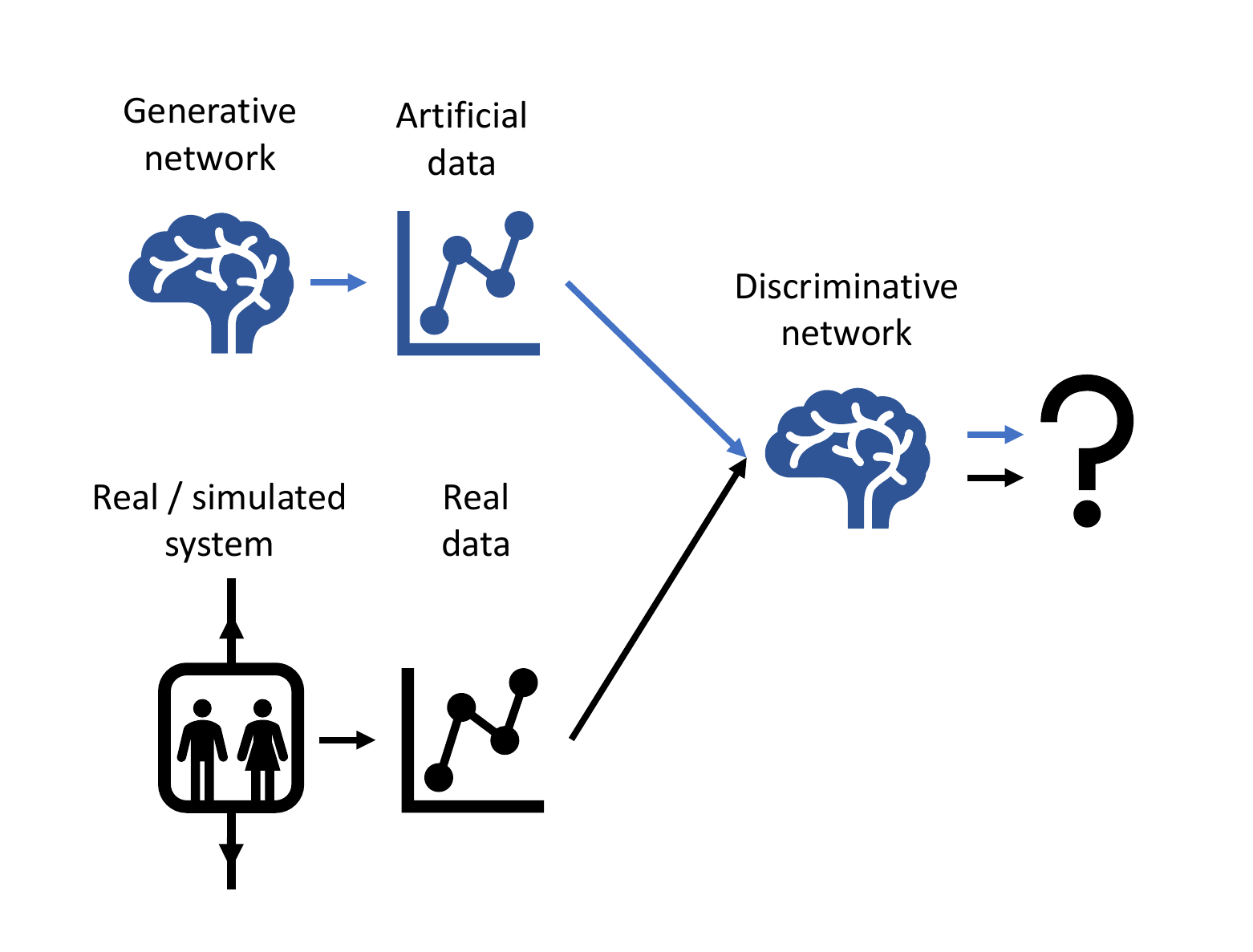}
    \caption{Architecture of a GAN. The discriminative network attempts
    to determine whether data is real or fake, whereas the generative network attempts to generate
    data that is not determined to be fake by the discriminative network.}
    \label{fig:gan_arch}
\end{figure}

\glspl{GAN} have been first proposed by \citet{goodfellow_generative_2014}. 
They describe a framework for generative modeling which consists of two separate networks, a generative network, $G$, and a discriminative network, $D$.
Both networks are trained in an adversarial process.
That means, the network $G$ tries to counterfeit data that deceives $D$ and $D$ is trained to discriminate between real training data and the `fake' data generated by $G$.
Figure~\ref{fig:gan_arch} illustrates this general idea.

This process results in both $G$ and $D$ improving their ability to fulfill their task, 
which in theory should create a generative network which creates 
data samples that are very hard to distinguish from real samples.

\glspl{GAN} have mostly been used to train generators for image data sets~\cite{goodfellow_generative_2014,yu_seqgan_2017}.  
Images are a well-suited application for \glspl{GAN}, because the generative network can make continuous adjustments to the generated data\cite{berges_text_nodate}.
This can easily be achieved for images, because a numerical color value might be changed by a fraction of 1\%.
This is not possible if discrete data such as text have to be processed. 
In this case, small changes may generate text sequences that have an entirely different meaning~\cite{yu_seqgan_2017}.

\citet{yu_seqgan_2017} propose a possible solution called \gls{sGAN} for this issue by adding feedback for unfinished sequences to the generator based on possible outcomes for a finished sequence.
As the discriminator is only able to calculate rewards for finished sequences, an additional step is taken by spanning a search tree for $n$ possible sequences based on the current time step and getting feedback for these from the discriminator.
The next step is then chosen from the best candidate. 
This process is repeated until a whole sequence has been generated.
Thereby, long-term rewards are taken into account for each time step instead of just focusing on short term rewards, resulting in a better quality of the generated sequences as a whole.
\gls{sGAN} showed promising results when tested on the generation of Chinese poems and political speeches, 
outperforming other approaches for sequence generation~\cite{yu_seqgan_2017}.
A more detailed description of \glspl{sGAN} can be found in Section~\ref{tools-seqgan}.

\section{Methods and Tools}\label{sec:methods}

\subsection{MceSim}

The training data for this project was generated using the \gls{mcesim}.
\gls{mcesim} is an open-source application for simulating multi-car elevator systems developed by~\citet{miyamoto_mcesim_2008}.
The simulation is designed for testing control patterns of elevator systems. 
It allows for different building configurations containing the amount of shafts in the building, cars per shaft, and the different types of floors in a building.
Internally, the simulation software processes the calls and prevents overlapping (colliding) cars.
Elevator calls can be created randomly by the simulation (as used in our experiments) or by providing a file that contains a list of the calls. 

The code of the simulator was modified in order to generate the output of the internal states when processing calls into a text file.
An example is shown in listing~\ref{lst:logfile}.
Each line consists of a timestamp (the internal time of the simulation) and an operation which is executed on a call.

\begin{lstlisting}[caption=MceSim Custom Log File,label={lst:logfile}]
114 - New call: call_1 from 4 to 1 guests 4
114 - Assign call: call_1 on car_01_01
120 - Load call: call_1 on car_01_01
131 - Unload call: call_1 on car_01_01 overtravel 0
\end{lstlisting}

During the simulation, many of these calls are handled in parallel.
The life cycle for each call is composed of the following operations, which always occur in this order:
\begin{description}
\item[New call] Passengers call an elevator inputting target location and the amount of passengers.
\item[Assign call] The call is assigned to a car.
\item[Load call] The car arrives at the start level and loads the passengers.
\item[Unload call] The car arrives at the destination level and unloads the passengers. 
\end{description}

To create a basic log file with entries that can be used as training data for the \gls{GAN}, a single log was generated with the following parameters:
\begin{lstlisting}[caption=basic log command,label={lst:logcmd}]
java -jar MceSim2018.jar -r -t 1000000      
\end{lstlisting}

The command specifies to use random calls and to run the simulation for 1e6 time units with the default configuration of five shafts, three cars each, and thirty floors.
This generates a log file with approximately 21,000 lines, each containing an operation on a call.
For the purposes of testing the approach, this single log was chosen as a simple dataset to test whether 
the application of \glspl{GAN} is feasible in the first place.
In addition to the simple dataset, a more complex dataset can be created 
by iterating through all provided configurations and running the simulation for 1e4 time units.

\subsection{GAN implementation}\label{tools-GANs}
\subsubsection{Tooling}\label{tools-software}

The programs used in our experiments were written in the Python programming language. The Keras framework as provided by Tensorflow was used to implement the neural networks. The following software versions were used:
\begin{itemize}
    \item Python 3.8
    \item Tensorflow 2.2.0 
    \item Numpy 1.19
\end{itemize}

Tensorflow and numpy can be installed using \texttt{pip} or by creating a \texttt{virtualenv}\footnote{see \url{https://virtualenv.pypa.io/en/latest/}}.
It is recommended to utilize GPU acceleration for training and inference, as especially the generative network is very resource and time consuming.
To enable GPU acceleration, tensorflow-gpu and and required drivers and libraries like Cuda and cuddn need to be installed\footnote{see for details: \url{https://www.tensorflow.org/install/gpu}}.

\subsubsection{Data Preprocessing}
Because neural networks internally use numerical values such as integers or floats, data usually must be converted before they can be passed to the network.
For images, this step is fairly simple because a 2D image can be represented as a 3D tensor of which the first two dimensions describe the position of a pixel while the third dimension describes its color.
For textual input, a dictionary needs to be defined. It is used to translate human readable text into its integer representation.
This was achieved by first creating the vocabulary \texttt{vocab} of the dataset containing each individual character.
After that, each vocabulary entry is mapped to an integer. 
The resulting dictionary \texttt{char2idx} is used to translate text to its integer representation, which can be processed by the network.

A character-based vocabulary was chosen because of the structure of the data.
This might appear counter-intuitively at the first sight, because only a few keywords are used in the log files.
However, as can be seen in listing~\ref{lst:logfile}, ascending IDs and timestamps result in increasingly large vocabularies, almost proportional to the length of the logfile.
A character-based logfile results in only 36 vocabulary entries or even only 32 when converting all letters to lower case.
Furthermore, the character based vocabulary is needed for the generator to be able to create new timestamps and IDs, which are not present in the training data and would as a consequence not appear in any vocabulary.

The dataset uses the format \texttt{[BATCH\_SIZE]}\texttt{[SEQUENCE\_LENGTH]}\texttt{[1]}.
Because of the variable length of the input text and the required uniform data shape, some remaining characters had to be cut off at the end of longer texts.
For data loading, shaping, and preprocessing, the resources from \citet{keras_engineer_intro}, \citet{deep_learning_python},  \citet{char_text_preprocess}, and  \citet{text_data_preprocessing_keras} were used.

\subsubsection{Recurrent Neural Networks}
\label{tools-rnn}
The first part of the \gls{GAN} is the generative network $G$.
Based on their performance for text generation and the the promising results of the \gls{sGAN} approach~\citep{yu_seqgan_2017}, \gls{RNN}s were chosen for the generator.
By utilizing \gls{LSTM}  layers, issues like the ``vanishing gradient problem'' or the ``exploding gradient problem''  are mitigated~\citep{yu_seqgan_2017}.
Vanishing gradients describe situations in which the model is not actually learning anything because the gradients are too small to have a effect. 
The exploding gradient problem describes the opposite issue where gradients are too large and result in very unstable network behaviors~\citep{pykes_vanishingexploding_2020}.
\glspl{LSTM} are commonly used to prevent this and are capable of handling large input sequences of varied length~\citep{yu_seqgan_2017}, which is required for our experiments.
The \gls{RNN} is created as a sequential model in Keras, containing the following layers:
\begin{itemize}
    \item Embedding layer, to break down high dimensional data into low dimensional data, to make it more understandable for the model.
    \item \gls{LSTM} layer as the recurrent layer with added long term memory, to enhance the network's capabilities for handling long sequences.
    \item Dense layer, which reduces the dimensions so that the output of the layer represents the probability of each item in the vocabulary for the next word.
\end{itemize}
The model was kept simple in order to keep the focus on the development of the total \gls{GAN} architecture.
A more sophisticated architecture might be useful, when more statistics about general \gls{GAN} performance are available and parameters of the individual networks need to be tuned.

For inference, a sequence can be fed into the network, which results in the network predicting the next character for the sequence in form of a tensor containing the probability for each individual entry of the vocabulary.
For example, if the network is trained to predict the next character in the alphabet and the sequence \textit{abcdefgh} is given as input, the network might produce an output tensor as shown in Table~\ref{tab:gen_out_tensor}.
\begin{table}
    \centering
    \caption{Exemplary output Tensor of the generator}
    \label{tab:gen_out_tensor}
    \begin{tabular}{|c|c|c|c|c|c|c|c|}
    \hline
        a & b & c & d & e & f & g & i\\
        \hline
        0 & 0 & 0.05 & 0 & 0 & 0.05 & 0 & 0.9\\
        \hline
    \end{tabular}
\end{table}
In practice, it is recommended to include some randomness in the selection process for the next word, 
because otherwise results might include loops of the most common vocabulary entries~\cite{tensorflow_text_gen}.
For this model, samples are drawn randomly from the weighted output distribution.

\subsubsection{Convolutional Neural Networks}
\label{tools-cnn}
The next important part of the \gls{GAN} is the discriminator, for which a \gls{CNN} was chosen.
Part of the selection process involved an analysis of different network types for text classification.
\citet{medium_text_classification} evaluated three different types of networks for text classification.
The results show a slight advantage of the chosen \gls{CNN} compared to a \gls{RNN} or a \gls{HAN}  when looking at the training and validation accuracy, with the \gls{HAN} coming close to or surpassing the \gls{CNN} in some instances.
The main benefit, why this network was chosen for the GAN, was the time factor.
The \gls{CNN} needs only a fraction of the time per epoch when compared to the \gls{RNN} and \gls{HAN}.
While these benefits might be subject to how each network was built, the \gls{CNN} was chosen with the option to replace it later if needed.

Similar to the \gls{RNN}, a sequential Keras Model was used for the \gls{CNN}.
This model implemented the following layers:
\begin{itemize}
    \item Embedding layer, to break down high dimensional data into low dimensional data, to make it more understandable for the model.
    \item 1D Convolutional layer, which uses a window of a certain size and calculates descriptive numbers for each possible window position in the dataset. This layer provides the name for the network.
    \item GlobalMaxPooling layer, which acts similar to the Convolutional layer, but takes maximum value from each window position. Here, `global' indicates that the maximum of the complete sample is used.
    \item Dense layer with ReLu (Rectified Linear Unit)~\cite{machine_learning_glossary} activation, which filters out negative values and sets them to zero. Dense layers, which are also known as fully connected layers, are \enquote{hidden layer[s] in which each node is connected to every node in the subsequent hidden layer.} \cite{machine_learning_glossary}
    \item Dropout layer to prevent overfitting
    \item Dense layer on sigmoid activation, to translate calculated predictions into probability numbers between 0 and 1
\end{itemize}

These layers were chosen because of their prominence in text classification in a \gls{CNN}.
Further evaluation of an improved network setup is needed.
A \gls{CNN} works as a classification network.
This means, that the network is being fed data with corresponding labels, in the case of this \gls{GAN} \enquote{real} for the data from the simulator and \enquote{fake} for the data from the discriminator.
The network then learns to classify each set of input data into one of these classes and tries to maximize accuracy and minimize loss.
A standalone \gls{CNN} needs multiple datasets for training, validation, and testing, with corresponding label datasets which must be in such a shape, that the first dimension is equal to the corresponding text dataset.
It is recommended to select the second dimension to be either \texttt{[1]} or \texttt{[amount of label classes]} when using one hot encoding~\cite{datacamp_cnn_tutorial}.

\subsubsection{Sequential GANs}\label{tools-seqgan}
\glspl{GAN} work efficiently on continuous data, but not on discrete data.
Especially long sequences of discrete data, such as text, are hard to process for \glspl{GAN}.
The main benefit of using \glspl{GAN} on continuous data can be found when analyzing how \glspl{GAN} operate.
\glspl{GAN} process training data for the discriminator and use approaching methods to minimize loss, with approaching steps sometimes being as small as one percent when looking at one single value.
\begin{figure}
    \centering
    \includegraphics[width=0.5\textwidth]{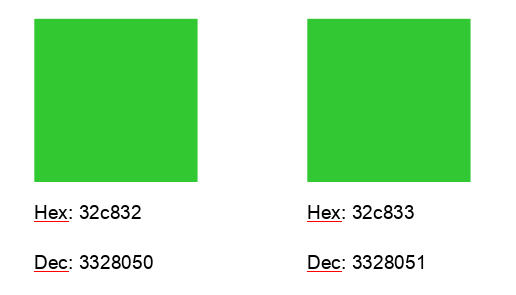}
    \caption{Small changes in a pixel result in not human perceptible change}
    \label{fig:change_color}
\end{figure}
\begin{figure}
    \centering
    \includegraphics[width=0.5\textwidth]{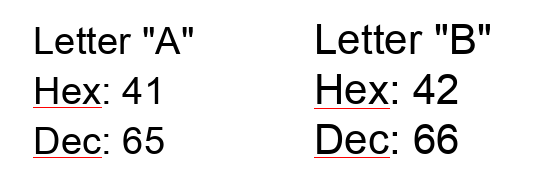}
    \caption{Small changes in ASCII values resulting in a changed meaning}
    \label{fig:change_ascii}
\end{figure}
Small changes like this can be made on continuous data like images without changing the depicted object or even without humanly visible changes at all.
Figure~\ref{fig:change_color} illustrates this effect.
When incrementing the hex value of the displayed pixel by just one, there is no visible change at all to the color.
However, when the same strategy is applied to discrete data like text, differences can get much more pronounced.
Figure~\ref{fig:change_ascii} illustrates how incrementing the ASCII value of the letter ``A'' by one results in ``B'' and therefore in a changed meaning.
Changes like this may have different consequences like spelling mistakes, creating completely unreadable sentences or even texts with completely different meanings.
For word based vocabularies, changes might even be more drastic, because slight changes might either change nothing at all (so the word remains the same) or a completely unrelated word might get chosen.

To tackle this issue, further research provided possible solutions.
The most promising solution found is the idea of using \glspl{sGAN} as proposed by~\citet{yu_seqgan_2017}.
They propose ``modeling the data generator as a stochastic policy in reinforcement learning'' \citep{yu_seqgan_2017}, meaning that the generator creates partial sequences, which then receive an expected result value, forcing the generator to learn and improve while building a dataset, instead of only optimizing the generator after finishing one dataset or epoch.

Due to the discriminator only being able to evaluate complete sequences, a strategy for evaluating incomplete sequences needed to be developed.
The \gls{sGAN} approach achieves this through building a Monte Carlo Search Tree describing $n$ possible complete sequences based on the current incomplete sequence.
Each possible next set of partial sequences is then discriminated by the discriminator, which allows for ranking the possible next parts of the sequences by their long term value.
To achieve a relatively accurate discrimination, the discriminator needs to be pre-trained with fake data from an untrained generator and real data.

The \gls{sGAN} algorithm can be described as follows:
\begin{enumerate}
    \item Import and shape data
    \item Build the network models (as described in Section \ref{tools-rnn} and \ref{tools-cnn})
    \item Pre-train the discriminator with training data
    \item Loop until the generator can reliably fool the discriminator
    \paragraph{Generator inner loop}
    \begin{enumerate}
        \item Generate a partial sequence
        \item Calculate the expected results for possible following sequences
        \item Optimize the generator with gradients drawn from the expected results
    \end{enumerate}
    \paragraph{Discriminator inner loop}
    \begin{enumerate}
        \item Discriminate real data and the fake date from the first inner loop
        \item Calculate losses
        \item Calculate gradients
        \item Train either discriminator or both networks
    \end{enumerate}
\end{enumerate}
A more detailed description can be found in \citet{yu_seqgan_2017}.

An implementation of this approach by Yu et. al. can be found on GitHub\footnote{See \url{https://github.com/LantaoYu/SeqGAN/}} \cite{yu_lantaoyuseqgan_2020} and was tested during this project, but proved to be outdated, using Python 2 and old Tensorflow libraries.
Additionally, no simple way to feed custom training data into the network was found without significantly modifying the implementation.

Considering the general focus of this project, it was decided not to use that implementation and instead focus on integrating the \gls{sGAN} approach into the \gls{GAN}  after finishing and evaluating it as a non-sequential \gls{GAN}.
This decision was also motivated by the fact that the \gls{sGAN} implementation uses Tensorflow directly without the Keras framework.

\section{Results}\label{sec:result}

The following subsections describe the results of standalone generator $G$ and discriminator $D$, 
and preliminary results of the \gls{GAN}.

\subsection{Standalone generator}
\label{result-gen}

In order to gather information about the performance of the generator $G$ without any feedback from the discriminator, the model was trained for ten epochs using the previously generated log files containing data of 1e6 simulation time units.
After that, the input ``\textit{1 - New Call:}'' was passed to the model, as each log starts with this sequence.
The model was used to generate 1e6 characters.
Generating the whole sequence took about 2.5 hours on a Nvidia RTX 2080Ti GPU, but the run time should be taken with some amount of uncertainty since the model is not optimized for performance and utilization of the GPU.
As a comparison, a similar amount of lines generated using \gls{mcesim} takes about ten minutes while only running on a single core of a CPU.
The first part of the resulting logfile is shown in Listing~\ref{lst:gen_text}.

\begin{lstlisting}[label=lst:gen_text,caption=Log generated by G]
1 - New call: c198196 from 19 to 27 guests 6
997953 - Assign call: c199257 on car_03_03
968675 - Load call: c192360 on car_02_01
979819 - Load call: c198112 on car_03_02
999762 - Unload call: c194291 on car_02_01 overtravel 0
993170 - New call: c198744 from 22 to 16 guests 8
979225 - Assign call: c197549 on car_04_02 overtravel 0
\end{lstlisting}

Some observations can be made in this output which can be generalized to the whole log:
\begin{itemize}
    \item The model was able to learn the general structure of the logfile. 
    This includes features like lines beginning with a timestamp followed by the type of operation on the call.
    Furthermore, the structure for each type of operation seems to be learned correctly.
    \item Timestamps are not ascending and most are in the high end of the range given by the input log file (near 1e6).
    Otherwise they seem to be random.
    \item IDs for calls and cars seem to be completely random.
    \item The ``life cycle'' for calls is not taken into account.
    Otherwise, all lines in the presented outtake of the log should start with \textit{new call} because IDs do not repeat.
\end{itemize}
Additionally, some incomplete lines like listing~\ref{lst:incomplete_line} (missing the additional information which car is assigned) or lines with random content like listing~\ref{lst:broken_line} can be found in the log.

\begin{lstlisting}[label=lst:incomplete_line,caption=Incomplete line,numbers=none]
994352 - Assign call: c1879902
\end{lstlisting}
\begin{lstlisting}[label=lst:broken_line,caption=Line with random content,numbers=none]
94_03
\end{lstlisting}

\subsection{Standalone Discriminator}
\label{result-dis}

\begin{figure}
   \centering
   \includegraphics[width=0.4\textwidth]{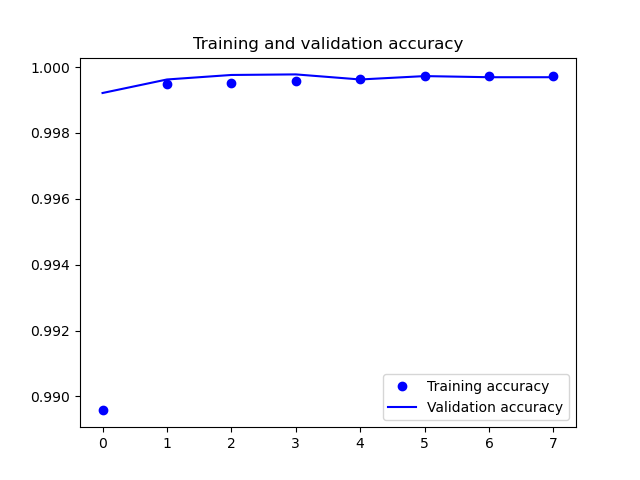}
   \caption{Standalone discriminator. Accuracy plotted against process steps. Larger values are better.}
   \label{fig:cnn_accuracy}
\end{figure}

\begin{figure}[h]
   \centering
   \includegraphics[width=0.4\textwidth]{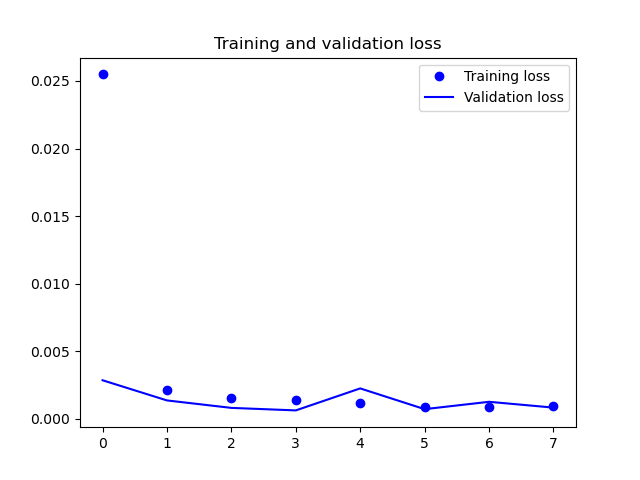}
   \caption{Standalone discriminator. Loss plotted against process steps. Smaller values are better.}
   \label{fig:cnn_loss}
\end{figure}

The discriminator as a standalone network was given \enquote{real} data from the simulator and \enquote{fake} data from the standalone generator as described in Chapter \ref{result-gen}.
The discriminator was quickly able to differentiate between real and fake data, which an accuracy of almost 99\% and a loss below 0.005 after the first epoch as seen in Figures \ref{fig:cnn_accuracy} and \ref{fig:cnn_loss} with slight improvements in the following epochs.

\subsection{GAN}\label{result-gan}

In the context of this project, a fully working implementation of the \gls{GAN} for text generation was not completed.
As described before, stand alone networks for the \gls{RNN} generator and the \gls{CNN} discriminator have been developed and tested.
When combining both, an issue for passing the data between generator and discriminator was discovered:
in the training loop, the generator generates a batch of predictions in the form of probability distributions as described in Section~\ref{tools-rnn}.
However, these cannot be used by the discriminator, because the discriminator expects an integer sequence as input.
As before, the solution is to sample from these batches to get actual text predictions, which in turn are now compatible to the discriminator.

Listing~\ref{lst:train_loop_bug} shows an excerpt from the training loop for the network.
The generator makes predictions for a complete batch, resulting in a tensor of the following shape:\\
 \texttt{[BATCH\_SIZE][SEQ\_LENGTH][VOCAB\_SIZE]}, representing the output distribution.
Next, a list for the predictions is created and a loop samples for each batch item, so actual sequences are obtained.
The resulting list of sequences is then transformed into an array of the shape \texttt{[BATCH\_SIZE],[SEQ\_LENGTH]} which can be passed to the discriminator.

\begin{lstlisting}[caption=Training loop,label={lst:train_loop_bug}]
batch_predictions = generator(input, training=False)

fake_list = []
for batch_prediction in batch_predictions:
  sampled_indices = tf.random.categorical(batch_prediction, num_samples=1)
  sampled_indices = tf.squeeze(sampled_indices,axis=-1).numpy()
  fake_list.append(sampled_indices)

fake_dataset = np.asarray(fake_list)
decision = discriminator(fake_dataset)
\end{lstlisting}

The process works fine for passing data between the two networks, but when the gradients of both networks are calculated, the gradients for the generator seem to be lost due to the conversion from the output tensor to an array.
Descriptions for this issue can be found in the project issues for Tensorflow\footnote{See \url{https://github.com/tensorflow/tensorflow/issues/36596}}.
In order to complete the implementation, a method for passing the data without losing gradients is needed.
Tests without the generator optimizer seem to indicate, that the network is otherwise functional.

Even though the  complete \gls{GAN} could not be implemented as planned, some recommendations from the literature will be presented: most authors note that \glspl{GAN} suffer from the issues presented in section~\ref{tools-seqgan}.
For example, \citet{berges_text_nodate} explored different models in a  \gls{GAN} setup and concluded that they were not able to create a generator which was able to deceive the discriminator.
Issues seem to be caused by the problem of sampling the probability distributions before passing sequences to the discriminator.

To alleviate these issues, the \gls{sGAN} approach proposed by \citet{yu_seqgan_2017} seems to be a promising solution. They present a way to give feedback to the generator while prioritizing the long term outcome for the generated sequence.
Their results show that the approach can be successfully used for natural language (political speeches and Chinese poems).

\section{Discussion}\label{sec:discussion}
The results of a standalone \gls{RNN} generator look promising.
It quickly learns the basic structure of the log file, the possible line types and correct spelling, even though it is trained on character basis.
However, other issues, like incorrect timestamps and random IDs, indicate that the model is not able to learn the broader contexts in the log file.

Moreover, the time, energy, and computation resources required to generate log files similar to those of the \gls{mcesim}, are much higher.
The example sequence of 1e6 characters requires on an GPU accelerated network about 15 times longer than the simulation utilizing only a single CPU core, making it much more expensive without any additional benefit in this concrete use case.

An evaluation of the discriminator is difficult at this point, without evaluating the generator at the same time.
It is possible that we observed rather good accuracy and loss values, because the discriminator learns fast and efficient, or else, because the generator creates text which is easy to detect as fake.
The discriminator would also have to be trained with more and a bigger variety of datasets to produce a more telling output.
As the difference between real and fake logs can be very easily determined by a human, it is assumed that the discriminator learns to check for similar features.
Unordered timestamps, no recurring IDs and incomplete lines might be artifacts which are picked up by the discriminator quickly.

Completing and testing the \gls{GAN} with a combination of both networks should give better insight into how the networks performed.
As described before, the implementation was not completed due to vanishing gradients  when sampling from the output distribution of the generator.
Testing the completed implementation can possibly give more insight about the usefulness of \glspl{GAN} for generating machine readable data like \gls{mcesim} log files.

Some observations made on the single tested networks are also true for \gls{GAN} in general.
\glspl{GAN} consume much more resources and power than the simulation software when only used to create more of the same data like in this project.
Additionally, the literature suggests in part, that the quality of the generated data might never be good enough to deceive the discriminator\cite{berges_text_nodate}, which needs to be tested with the completed implementation.
Because of the simple structure of the training data compared to natural language, results for this application might be more favorable.
Additions like the \gls{sGAN} approach~\citep{yu_seqgan_2017} might improve on this and could be adapted to the simpler structure of machine readable data. 

\section{Outlook}\label{sec:outlook}

This section presents suggestions for future work.
\begin{description}

\item{Finishing the GAN implementation:}
\label{future_finish_gan}
Most importantly, the implementation of the \gls{GAN} prototype should be finished to enable an extensive performance evaluation. 
While the literature suggests that \glspl{GAN} are not necessarily a good approach for generating text data, there are approaches which seem to be promising.
Additionally, the data handled in this project is of much simpler structure than natural language which is usually considered when talking about the difficulties of text processing.

\item{Improving the networks:}
\label{furute_improve_gan}
After finishing and evaluating the current \gls{GAN} configuration, improving the networks might be the next step.
\citet{lucena_structureboost_2020} suggested using \glspl{SCDT} instead of numerical or one hot encoding for improving Gradient calculation.
This might accelerate learning for the networks.
An evaluation of \glspl{SCDT} for \glspl{GAN} might provide some benefits regarding learning speed.

As a majority of the available time was consumed understanding and building the networks and combining them into the \gls{GAN}, no further evaluation of specific network layers or combination of layers was made.
With a more thorough understanding of neural networks and their corresponding layers, their parameters and combination and order of layers, the generator and discriminator could be improved.
Further testing and evaluation might even suggest using different networks for the generator and discriminator instead of the \gls{RNN} and \gls{CNN} that were used in our experiments.

\item{Evaluating \glspl{GAN} and output data:}
\label{future_evaluating}
Another issue that should be investigated in subsequent studies is the evaluation of the \gls{GAN}, the finished output networks, and the generated data.
While this might give suggestions on how to improve the networks, it is also important to evaluate before using the finished generator or generated data in critical fields such as elevator planning, vehicle programming or medical analysis.
Using faulty or inaccurate data might lead to failures that could affect human safety, health or lives.
While no solution to evaluate data for safe usage in critical fields was found during this project, the research of \citet{esteban_real-valued_2017} suggests two ways of evaluating and optimizing \glspl{GAN}.
The first approach is to check whether the \gls{GAN} \enquote{implicitly learns the distribution of the true data} \cite{esteban_real-valued_2017} via a medium mean discrepancy, while the second approach suggests training a second  \gls{GAN} with the generated data of the first  \gls{GAN} and testing it with real data.
If the second  \gls{GAN} is trained with generated data from the first \gls{GAN}, it can successfully detect real data as real, then we can consider the first  \gls{GAN} a success.

\item{Implementing a sequential GAN:}
\label{future_seqgan}
Another possible addition after finishing the  \gls{GAN} implementation may be the  \gls{sGAN} approach as described in section~\ref{tools-seqgan}, which is expected  to work better on discrete data such as text compared to a more standard  \gls{GAN} design.
Since the architecture of the \gls{sGAN} foresees an additional search algorithm, as well as having the discriminator calculate losses for multiple partial sequences per sequence generated, the resource cost concerning CPU, GPU and time usage is expected to increase.
It is to be evaluated if the benefits outweigh the additional cost or if similar results can be achieved by using the non-sequential \gls{GAN}.
An additional idea might be to evaluate the used search algorithm.
Yu et. al. use a Monte Carlo Search Tree, but there might be more cost efficient alternatives for this scenario.

\item{Adding building configuration input into the GAN:}\label{future_building_conf_input}
To enhance the usefulness of a complete application around the  \gls{GAN}, it would be helpful to be able to pass a building configuration describing the amount of floors, shafts, and cars per shaft.
The  \gls{GAN} needs to be able to process this input and generate logs which apply to exactly this building configuration.
The feature is required to create data which can be used for optimization of elevator systems in specific scenarios. 
This might be achievable by generating datasets containing only logs for the appropriate building and training the network with this specific dataset.
To reduce training times, it might be appropriate to pre-train a general model. This can be adapted using problem specific data, if available.
\end{description}

\section{Conclusion}\label{sec:conclusion}
This article investigates the applicability of GANs for imitating complex and expensive real-world simulations. 
We compared the simulation output of a technical system with the output of a GAN. 
To exemplify this approach, a well-known multi-car elevator system simulator was chosen. Our study demonstrates the feasibility of this approach. It also discusses pitfalls and technical problems that occurred during the implementation. Although we were able to show that in principle, GANs can be used as substitutes for expensive simulation runs, we also show that they cannot be used ``out of the box''. Fine tuning is needed. 
We presented a proof-of-concept, which can serve as a starting point for further research. 

Overall, our research provided much insight in the application of \glspl{GAN} for the generation of machine-readable data, 
even though a working implementation could not be finished. 
\glspl{GAN} are  powerful tools for creating data, but also come with many difficulties when working on discrete data.
Some of these issues are due to the conceptual design of models that were created with image manipulation in mind, while others are more technical like the gradient issue for the implementation of this project.

Applying the \gls{GAN} framework to text files, which can also be generated using the actual simulation itself, requires more resources and time.
Especially, the generative network takes large amounts of training data and time to be able to generate data which is still easily distinguishable from real training data, by human eye as well as the discriminative network.
However, many improvements of the individual networks and the overall \gls{GAN} can be made to optimize the quality of the outcome.
Building more sophisticated network models and adding concepts like the sequential approach might improve the output of the generative network sufficiently. Then, logs could mimic those of the simulation or even formats like those produced by real world elevator systems.

Even though this approach may seem cost- and performance ineffective at the first glimpse, there might be some additional benefits when looking beyond the scope of this project, like the creation of real world like data which must not include real details, e.g., for privacy protection in medicine or enhancing real-world datasets generated by actual elevator systems.

\bibliographystyle{plainnat} 
\bibliography{references}

\end{document}